\documentclass[10pt,twocolumn,letterpaper]{article}

\usepackage{3dv}
\usepackage{times}
\usepackage{epsfig}
\usepackage{graphicx}
\usepackage{amsmath}
\usepackage{amssymb}

\usepackage[english]{babel}
\usepackage[utf8]{inputenc}
\newcommand\figref {Figure~\ref}
\usepackage{subcaption}
\usepackage[font = rm]{caption}
\usepackage{enumitem}
\usepackage{subfiles}
\usepackage[pagebackref=False,breaklinks=true,letterpaper=True,colorlinks,bookmarks=false]{hyperref}

\threedvfinalcopy 

\def\threedvPaperID{95} 

\ifthreedvfinal\fi
\begin{document}

\title{GASCN: Graph Attention Shape Completion Network}

\author{Haojie Huang\qquad Ziyi Yang\qquad Robert Platt\\
Northeastern University\\
360 Huntington Avenue, Boston, MA 02115, United States\\
{\tt\small \{huang.haoj; yang.ziyi2\} @northeastern.edu; rplatt@ccs.neu.edu}
}

\maketitle
\thispagestyle{empty}

\begin{abstract}

Shape completion, the problem of inferring the complete geometry of an object given a partial point cloud, is an important problem in robotics and computer vision. This paper proposes the Graph Attention Shape Completion Network (GASCN), a novel neural network model that solves this problem. This model combines a graph-based model for encoding local point cloud information with an MLP-based architecture for encoding global information. For each completed point, our model infers the normal and extent of the local surface patch which is used to produce dense yet precise shape completions. We report experiments that demonstrate that GASCN outperforms standard shape completion methods on a standard benchmark drawn from the Shapenet dataset.

   
\end{abstract}


\section{Introduction}

It is often desirable to be able to infer the geometry of objects in a scene based on a small number of sensor measurements. This is known as the \textit{shape completion} problem where the system infers the complete geometry of an object based on a partially observed point cloud, such as that produced by a depth sensor. In particular, shape completion could be useful in the context of robotic manipulation where a robotic system could use the inferred geometry of the objects in a scene to create geometric plans that solve desired tasks. Compared to other approaches to scene reconstruction that estimate the pose of specific objects, a critical advantage of shape completion methods is that they can be used to infer the geometry of novel objects, not just a closed set of previously modeled objects.

There are two main types of approaches to shape completion: one based on the point cloud representation and the second based on voxel grids or signed distance functions. Point clouds are a convenient choice because they can be created easily from the output of depth sensors or LIDAR and several methods use point cloud represenations~\cite{yuan2018pcn,Yang_2018_CVPR,groueix2018papier,liu2020morphing}. However, these methods are all limited by the fact that they cannot use 3D convolutions because the point cloud representation does not explicitly encode information about the local connectivity of each point. An alternative is to create shape completion algorithms based on a voxel grid or signed distance function (SDF) representation~\cite{sarmad2019rl,han2017high,stutz2018learning}. Here, we can use 3D convolutions to obtain higher shape completion accuracy, as in GRNet~\cite{xie2020grnet}. However, the memory requirements for voxel grids or SDFs are cubic in the resolution of the grid and therefore it severely limits the resolution at which the method can be applied.


In contrast to the methods above, this paper proposes GASCN, a shape completion approach that operates on a graph rather than a point cloud or voxel grid. Since converting a point cloud into a mesh can itself be challenging, we use a graph attention approach~\cite{velivckovic2017graph} where each point in the cloud is associated with a unique graph that connects it with its neighbors. Another key element of our approach is a densification step that converts a coarse reconstruction into a fine reconstruction by deforming a fine 2D grid to each coarse point via a learned surface normal and variance. The novel features of our approach relative to prior shape completion work are the following:

\begin{enumerate}
\setlength{\itemsep}{10pt}
\setlength{\parsep}{2pt}
\setlength{\parskip}{2pt}
\item Our encoder utilizes the graph attention layer to encode local context information for further combination with the global structure information whereas prior work on shape completion typically uses a PointNet~\cite{Qi_2017_CVPR} encoder or 3D convolutions on voxels~\cite{wu20153d}. This enables us to take point cloud input while still encoding using graph convolutions.

\item Our decoder contains a novel densification procedure that maps a dense planar grid to a local region around each point in a coarse shape completion. This enables us to produce a dense completion while keeping the network complexity manageable. 

\end{enumerate}

We compare our model with several other recent approaches from the literature~\cite{Yang_2018_CVPR,yuan2018pcn,Tchapmi_2019_CVPR} and find that it outperforms on a standard shape completion baseline task.

\vspace{-0.2cm}
\section{Related Work}

The problem of inferring 3D structures from RGB and depth camera images has been studied for decades. A classic but powerful non-learning method uses the Iterative Closest Point (ICP) ~\cite{besl1992method} algorithm to align a template model from a large dataset with the observed point cloud.
BEOs~\cite{burchfiel2017bayesian} constructs a new representation space with VBPCA~\cite{bishop2006pattern} basis, where the best corresponding point to the partial input could be found.
Some other works~\cite{rock2015completing,Gupta_2015_CVPR,li2016shape} deform the template model to synthesize shapes that are more consistent with the observations. These methods are often sensitive to noise and require large datasets as well as expensive computation time.

Recent learning-based approaches complete the 3D structure with a parameterized model, often a deep neural network, directly from the observation. It allows a fast inference and great generalization. Some datasets like ShapeNet~\cite{shapenet2015} and YCB~\cite{calli2017yale} enable learning on sufficient examples to generate visually compelling shapes. The most common structure for a shape completion network consists of an encoder that maps the partial observation to a latent vector and a decoder that produces the complete 3D shape given the latent vector. The learning method differs based on the different representation methods of a 3D object. 

Voxel-based methods~\cite{stutz2018learning,wu20153d,han2017high} voxelize point clouds into binary voxels, where 3D convolutional neural networks can be applied. Recently, GRnet~\cite{xie2020grnet} proposed a novel gridding mechanism to encode the neighbors' information into the vertices of each voxel. These methods are often computation expensive and limited to quantization effects due to the voxel representation. 

PointNet~\cite{Qi_2017_CVPR} is a pioneer in using the MLP-based approach to learn point clouds. It combined pointwise multi-layer perceptrons with a max-pooling aggregation function to achieve invariance to permutation and robustness to perturbation. After PointNet, FoldingNet~\cite{Yang_2018_CVPR} proposed a folding-based decoder that deforms a 2D grid onto the 3D surface of a point cloud. Following previous works, PCN~\cite{yuan2018pcn} proposed the coarse-to-fine procedure to densify the point cloud. AtlasNet~\cite{groueix2018papier} and MSN~\cite{liu2020morphing} generated a collection of parametric surface elements based on FoldingNet's decoder. Besides, some work~\cite{li2018point,wu2016learning} combined MLPs with GANs~\cite{goodfellow2014generative}, and some~\cite{sarmad2019rl} integrated it with reinforcement learning. However, these approaches just aggregated the global information from the whole point cloud without focusing on the local information of each point's neighbors, and neither did they take the various local properties of different points into account during the generation process. 

Compared with MLP-based approaches, graph-based methods could extract local information of each point. Most recent graph-based convolution methods~\cite{kipf2016semi,hamilton2017inductive,velivckovic2017graph} operated on groups of spatially close neighbors. ECC~\cite{simonovsky2017dynamic} is the first one to apply graph convolutions to point
cloud classification. After that, DGCNN~\cite{wang2019dynamic} proposed a more generalized edge convolution method with dynamic graph updates to encode the local neighborhood information for point clouds. Inspired by DGCNN, DCG~\cite{wang2019deep} introduced this method to point cloud completion following a coarse-to-fine fashion. However, the edge convolution is computationally expensive in large graphs. After Graph Attention Network~\cite{velivckovic2017graph}, GAC~\cite{wang2019graph} inherited its ideas and applied attention mechanism to the point cloud segmentation area.
To the best of our knowledge, we are the first to directly combine the graph-level local information with the global structure information in the encoder as well as utilize surface normals to incorporate pointwise local properties for shape completion.

\section{Method}
\label{chap:method}

\begin{figure}[htbp]
    \centering
    \includegraphics[width=0.45\textwidth]{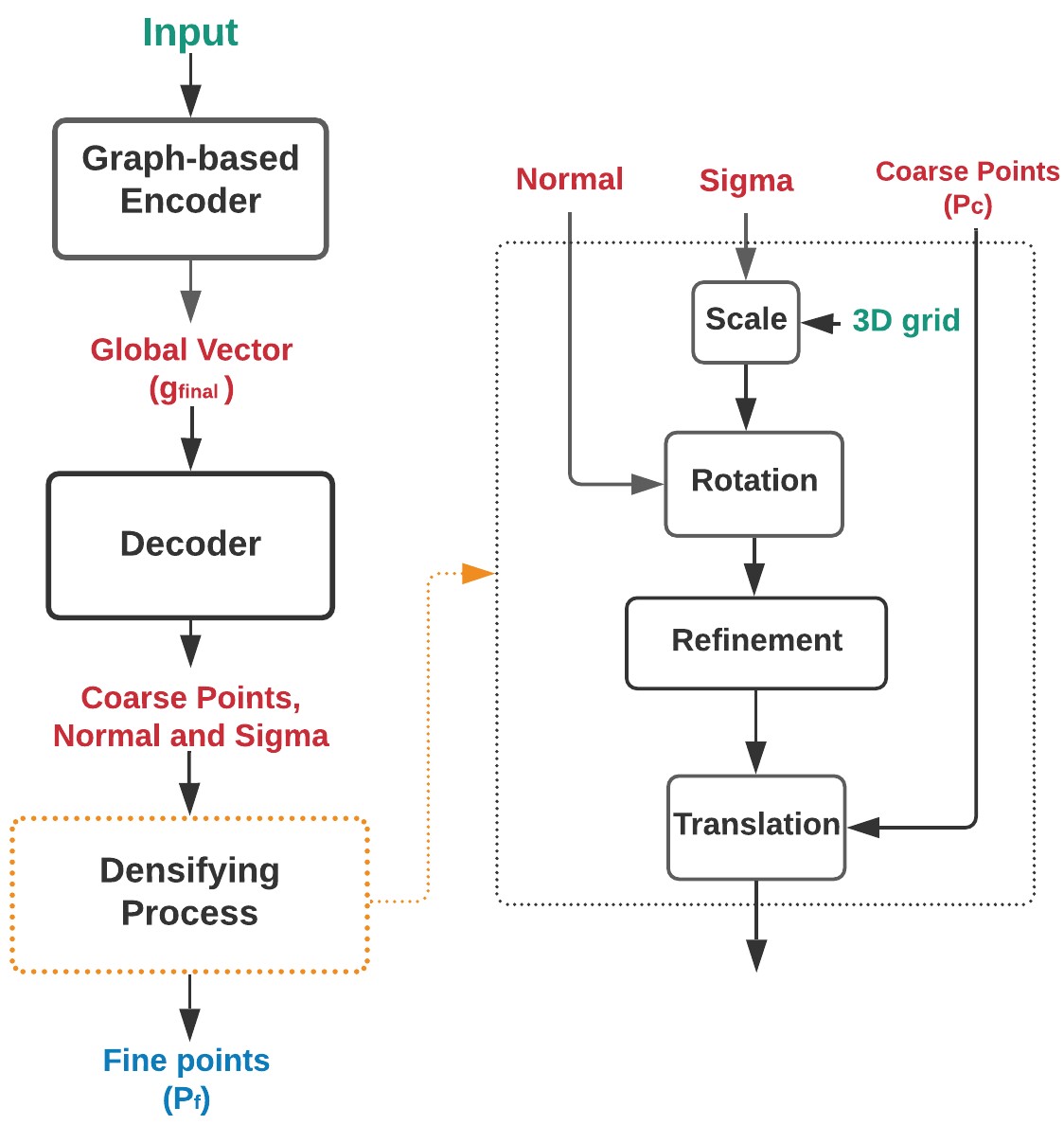}
    \caption{The architecture of GASCN. It has a graph-based encoder that integrates each point's local information with the point cloud's global structure information. The decoding process utilizes surface normals and coarse points to rotate and translate adaptive 3D grids to densify coarse point clouds.}
    \label{fig:method:network}
\end{figure}

\figref{fig:method:network} below is an overview of our network, which follows an encoder-decoder architecture. It takes a partial point cloud with a variable number of points as input, $p_{input}$ and generates a complete dense point cloud as output, $p_{final}$.

\begin{figure}[htbp]
    \centering
    \includegraphics[width=0.5\textwidth]{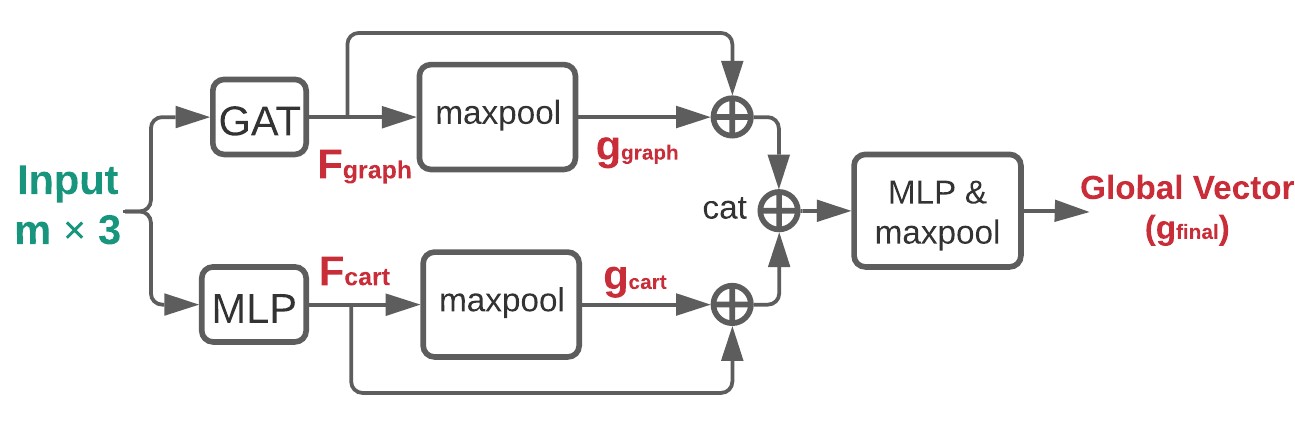}
    \caption{Graph-based Encoder of GASCN model. It encodes each node feature with its neighbors' information $F_{graph}$ by graph attention layer, its mapped Cartesian feature $F_{cart}$ by point-wise MLP, and two types of global vectors, $g_{graph}$ and $g_{cart}$, extracted from the max-pooling operation.}
    \label{fig:method:encoder}
\end{figure}

The encoder (\figref{fig:method:encoder}) uses a graph attention network (GAT)~\cite{velivckovic2017graph} to reason about the local geometry of the input cloud and MLPs to reason about global geometry. 

GAT~\cite{velivckovic2017graph} is used to learn the graph-level local information and MLPs are used to encode the global information. Following a coarse-to-dense procedure, our decoder first decodes a sparse point cloud $P_c$ and then samples each point's neighbors to densify $P_c$.
To make the densifying process flexible and adaptive to different local geometries, we decode surface normals and the scale factor $\sigma$ by our normal decoder and sigma decoder to adjust the orientation and the size of each point's neighborhood.

\subsection{Encoder}
Since each region of a point cloud could have different geometries, all the global information without the local information would result in averaging over all point features and thus undermining the completions. In order to solve it,
our encoder is built on the graph representation of partial point clouds. We encode each node feature with its neighbors' information, its mapped Cartesian feature, and two different types of global vectors to maximize the utilization of local information and global information.

\subsubsection{Build Graph for Point Cloud}
\label{chap:intro:design}
Given a point cloud consisting of $m$ points, denoted as $P = \{p_1,p_2,...,p_m\}$ with $p_i\in\mathbb{R}^3$, we form a directed graph $G = (V, E)$ representing the point cloud structure, where $V=\{1,...,m\}$ and $E \subseteq V\times V$. We define $H = \{h_1,h_2,...,h_m\}$ with $h_i \in \mathbb{R}^D$ as a set of node features corresponding to each point, where $D$ is the dimension of the node features. Each node feature is initialized with ${h_i}=(x_i,y_i,z_i)$, where $(x_i,y_i,z_i)$ is the Cartesian coordinates of point $p_i$. 
Moreover, we build subgraphs $\{G_i\}_{i=1,2,...,m}\subseteq G$ for all nodes as k-nearest neighbor~\cite{peterson2009k} graphs from node features ${H}$ by connecting the corresponding point $p_i$ with its k-nearest spatial neighbors. Besides, the subgraphs include self-loops to preserve each node's own information.

As a result, the whole graph for the point cloud comprises $m$ directed subgraphs and each point is assigned as the destination node of a subgraph. Each of them seems like a patch on the surface of a 3D object, similar to the kernel window in 2D convolution. 

\subsubsection{Local Convolution: GAT}
\label{chap:intro:design}

Although points in the point cloud are independent, there must exist certain hidden relationships to maintain the local surface geometry. GAT~\cite{velivckovic2017graph} provides an attentive mechanism to learn it.
Given the graph $G$ and point features ${H}$ of a partial point cloud, we apply the GAT layer to aggregate each node's neighborhood information.

First, we conduct a linear transformation for the embedding feature ${h_i}$ of the node $i$ with a shared learnable weight matrix $W$ and then compute a pair-wise unnormalized attention score $e_{ij}$ with a learnable vector $a$ between the two neighbors $i$ and $j$ ($||$ is the concatenation operation).

\begin{equation}\label{eq1}
    z_i = Wh_i
\end{equation}
\begin{equation}\label{eq2}
    e_{ij} = LeakyReLU(a^T(z_i||z_j))
\end{equation}

After that, our network normalizes the attention score on each node's neighbors and aggregates the embeddings from neighbors with the normalized attention score, and applies a nonlinear activation function.

\begin{equation}\label{eq3}
    \alpha_{ij} = \frac{exp(e_{ij})}{\sum_{k\in N(i)}exp(e_{ik})}
\end{equation}

\begin{equation}\label{eq4}
    h^{new}_{i} = LeakyReLU(\sum_{j\in N(i)}\alpha_{ij}z_j)
\end{equation}

The final result is denoted as $F_{graph}$ which is an $m\:\times\:d$ matrix, where $m$ is the number of points and $d$ is the feature dimension.
The process above maps each node feature from a 3-dimension $xyz$ coordinate to a $d$-dimension vector that contains the information of its neighbors through the graph attention layer.

\subsubsection{Global Information}
\label{chap:intro:design}
The global structure information is critical in point cloud generation, which could be used to distinguish different objects. We extract two different types of global information, the global graph vector $g_{graph}$ and the global Cartesian vector $g_{cart}$.
The global graph vector $g_{graph}$ is given by
\begin{equation}
    g_{graph} = maxpool(F_{graph})
\end{equation}
where max-pooling operation concatenates the maximum value among each feature dimension of $F_{graph}$ to construct the permutation-invariant global vector $g_{graph}$.

We define the pointwise MLPs in PointNet~\cite{Qi_2017_CVPR} as $M_{\theta}:\mathbb{R}^{d_1}\,\to\,\mathbb{R}^{d_2}$. It maps each point feature from $d_1$-dimension into a new $d_2$-dimension space using a nonlinear function $M$ with the shared learnable weights $\theta$.
The second global vector is aggregated by:
\begin{equation}
    F_{cart} = M_{\theta_1}(H)
\end{equation}
\begin{equation}
    g_{cart} = maxpool(F_{cart})
\end{equation}

After that, two different feature matrices, $F_{graph}$ and $F_{cart}$, are concatenated into an augmented feature matrix $\Bar{F}$, while two global vectors $g_{graph}$ and $g_{cart}$ are concatenated together to produce the augmented global vector $\Bar{g}$. We concatenate $\Bar{g}$ to each row of $\Bar{F}$ to get our final feature matrix $F_{final}$.
So far, each node feature is encoded with information of its neighbors, its mapped Cartesian feature, and two types of global vectors. Finally, we extract the final latent vector $g_{final}$ by:
\begin{equation}
    g_{final} = maxpool(M_{\theta_2}(F_{final}))
\end{equation}

\subsection{Decoder}
\begin{figure}[htbp]
    \centering
    \includegraphics[width=0.3\textwidth]{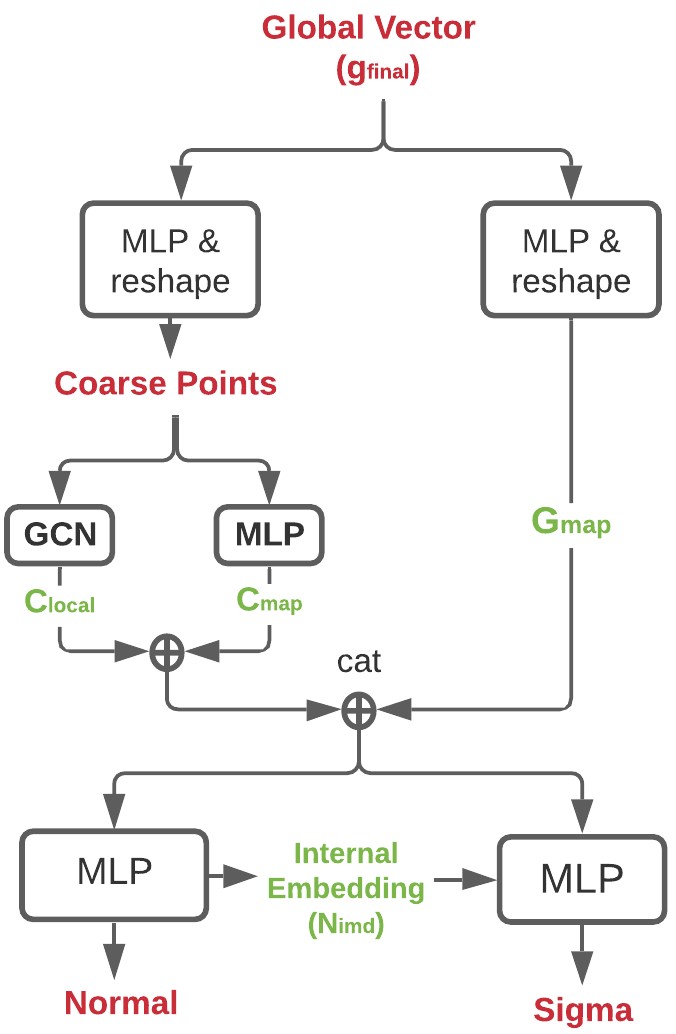}
    \caption{Decoder Architecture. The coarse points decoder consists of three fully connected layers. The normal decoder takes as inputs the mapped coarse point coordinate $C_{map}$, its neighbors' information $C_{local}$, and the mapped global vector $G_{map}$. The sigma decoder takes as inputs the normal internal embedding $N_{imd}$, $C_{local}$, $C_{map}$, and $G_{map}$, and then applies three pointwise MLP layers to output pointwise-attentive sigmas.}
    
    \label{decoder}
\end{figure}

Regarding the various local properties of different points during the generation process, our decoder first generates coarse points, each coarse point's surface normal, and its corresponding scale factor $\sigma$, as shown in \figref{decoder}. It then samples coarse points' neighbors with adaptive 3D grid points and utilizes both surface normals and coarse points to determine the rotation and the translation, respectively.

\subsubsection{Coarse Point Cloud and Surface Normal}

The coarse output $P_c$ is a preliminary generated complete shape with sparse points. Following previous work~\cite{yuan2018pcn}, we use three fully connected layers with nonlinear activation functions to decode the $ g_{final}$ to $P_c$, as shown at the top of \figref{decoder}.

In order to densify completions, some pointwise attributes are needed to generate neighbors.  Surface normal is a good option which contains the local orientation of a surface.
To decode surface normals, we first build a graph by connecting each coarse point with its nearest neighbors and apply one graph convolution layer~\cite{kipf2016semi} to aggregate the neighbor information into $C_{local}$.
In the meantime, the global vector $g_{final}$ and coarse points $P_c$ are mapped respectively by two different MLPs into $G_{map}$ and $C_{map}$. The surface normal is generated by

\begin{equation}
    N_{imd} = M_{\phi_1}(C_{local}\,||\,C_{map}\,||\,G_{map})
\end{equation}

\begin{equation}
    Normal = \frac{M_{\phi_2}(N_{imd})}{||M_{\phi_2}(N_{imd})||}
\end{equation}
where the $N_{imd}$ is the normal internal embedding. The whole process above is shown in \figref{decoder}.

\subsubsection{Calculate Neighbors by Surface Normals in Coarse Point Cloud}
\label{chap:intro:neighbors}

Given a point $\mathbf{p}=(p_x,p_y,p_z)$ and its normal $\mathbf{n}=(n_x,n_y,n_z)$, how could we sample its neighbor points? Neither sampling $X,Y,Z$ randomly ~\cite{liu2020morphing} from a distribution that would make the output look noisy and the model hard to train, nor utilize a fixed grid which is inflexible; we instead mesh an adaptive grid to obtain a set of 3D Cartesian coordinates for each coarse point.
Our procedure has four steps: sampling, scaling, rotation, and translation.

\begin{enumerate}
\setlength{\itemsep}{1pt}
\setlength{\parsep}{1pt}
\setlength{\parskip}{1pt}
\item  Obtain neighbors by meshing grid:

\begin{equation}
(X,Y) = Meshgrid(l, l);\:Z = \mathbf{0}
\end{equation}

$(X,Y,Z)$ is a set of vertices of a square grid of size $l$, centered at the origin with surface normal oriented along Z-axis, forming a plane in 3D space.

\item Scale $X,Y$ with $\sigma$:
\begin{equation}
x = \sigma\cdot X; y = \sigma\cdot Y
\end{equation}

Each point has a specific scale factor $\sigma$ that is learned from the neural network, which we will discuss in Section \ref{chap:method:sigma}. After the scaling, different points are associated with planes of different sizes accordingly.

\begin{figure}[htbp]
    \centering
    \includegraphics[width=0.4\textwidth]{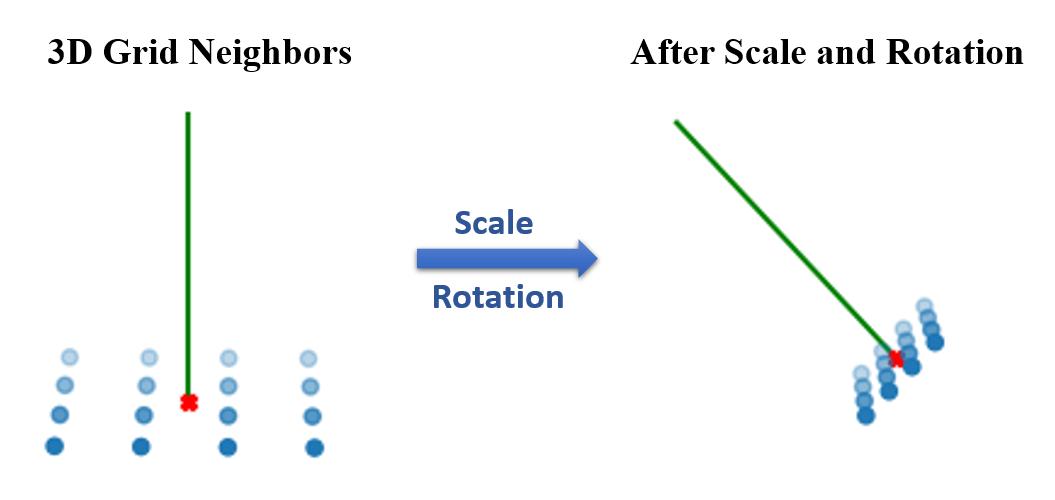}
    \caption[Neighbors]{
    After the scale and rotation, we align the initial 3D-grid neighbors' surface normal along the decoded $Normal$. This process enables our generated grid points to match local surfaces with different orientations and geometries. (Blue dots are neighbors, the green line indicates the normal direction, and the red dot is the center.)}
    \label{fig:method:rotation}
\end{figure}

\item Rotate the samples to make their surface normals align along the surface normals we generated, as shown in \figref{fig:method:rotation}:

\begin{equation}
    k = z_{axis}\times n
\end{equation}
\vspace{-0.3cm}
\begin{equation}
    \theta = \arccos{(n_z)}
\end{equation}

where the $k$ is the rotation axis, $z_{axis}$ is the unit vector along $Z$-axis, $\times$ is the cross product, $\theta$ is the rotation angle. If $k$ is a zero vector, no rotation is needed. We use Rodrigues' Rotation Formula to transform the axis-angle representation to a rotation matrix $R$.

\item Translate the center of the samples to point $\mathbf{p}$. The samples are centered at the origin $(0,0,0)$ previously, and we add the generated point's coordinate $(p_x,p_y,p_z)$ to each sample's coordinate to complete the translation.

\end{enumerate}

After applying the operations to each point of $P_c$, we fuse each point's local surface information with its coordinate to densify the point cloud.

\subsubsection{Scale Factor $\sigma$}
\label{chap:method:sigma}
$\sigma$ is the scale factor learned from our sigma network shown in \figref{decoder} to scale the samples. For each point in $P_c$, there is a $\sigma$ corresponding to it. It makes the neighbor generation be flexible and attentive. Using a table as an instance, the corner points or edge points may have smaller $\sigma$, while points on a flat plane should have a larger $\sigma$ since they have more confidence in sampling neighbors with a wider grid. In our work, $\sigma$ is pointwise attentive and learned by
\begin{equation}
    \log{\sigma} = M_{\phi_\sigma}(N_{imd}\,||\,C_{local}\,||\,C_{map}\,||\,G_{map})
\end{equation}


\begin{table*}
\centering
\begin{tabular}{c| cccccccc |c}
\hline
Model & Airplane & Cabinet & Car & Chair & Lamp & Sofa & Table & Watercraft & Overall\\
\hline
FoldingNet & 10.74 & 19.31 & 21.03 & 21.89 & 19.93 & 19.82 & 22.55 & 18.31 & 19.20\\
GRnet & 9.16 & 17.98 & 14.96 & 12.34 & 10.66 & 14.56 & 13.84 & 11.05 & 13.06 \\
TopNet & 7.12 & 14.54 & 13.19 & 13.65 & 11.51 & 12.80 & 12.88 & 10.13 & 12.05\\
PCN & 6.51 & 14.43 & 11.81 & 10.90 & 10.32 & 10.92 & 14.58 & 8.96 & 11.05\\
Ours & \textbf{4.71} & \textbf{11.69} & \textbf{9.67} & \textbf{8.44} & \textbf{6.91} & \textbf{8.31} & \textbf{10.02} & \textbf{6.69} & \textbf{8.31}\\
\hline
\end{tabular}
\caption{Point completion results \textcolor{black}{comparison} on ShapeNet using Chamfer Distance
(CD) with L2 norm computed on 16,384 points and multiplied by $10^3$. The best results are highlighted in bold.}
\label{tab:tab1} 
\end{table*}

\begin{figure*}
    \centering
    \includegraphics[width=0.9\textwidth]{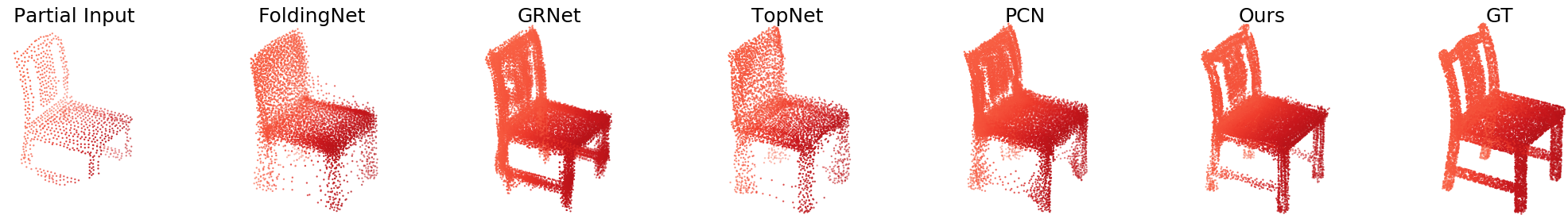}
    \includegraphics[width=0.9\textwidth]{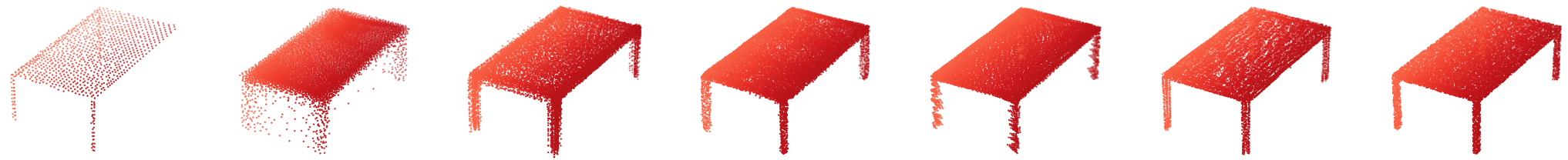}
    \includegraphics[width=0.85\textwidth]{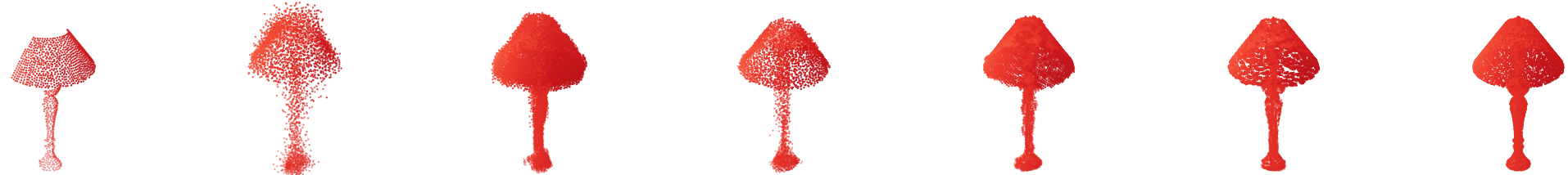}
    \includegraphics[width=0.9\textwidth]{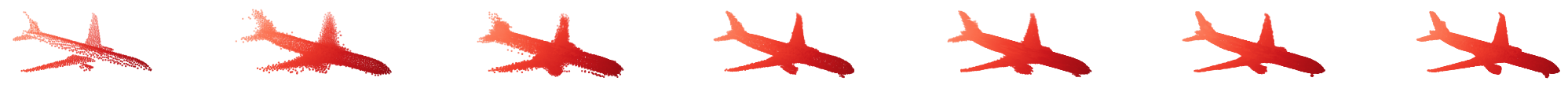}
    \caption{Visual Results Comparison\textcolor{red}{}. A partial point cloud is given and our method generates better complete point clouds.}
    \label{fig:exp:compare}
\end{figure*}


\subsubsection{Deformation and Refinement}
So far, our network generates a dense set of points, $P_d$. In order to deform the 3D-grid points and ensure robust performance, each point is concatenated with the corresponding coarse point, which serves as its positional embedding. We then map the feature to a high-dimension space and project them back to 3D space by three points-wise MLPs and Leaky-Relu activation function~\cite{xu2015empirical}.
We intuitively denote the improved dense point clouds as fine point clouds, $P_f$, which marks the end of our forward propagation.

\label{chap:intro:design}

\subsection{Loss}
\label{chap:intro:design}
Since the point cloud is an unordered set, we use the permutation-invariance loss, Chamfer Distance (CD) ~\cite{fan2017point}, to measure the distance between our output and the ground truth.
CD (\ref{eq:cd}) calculates the average closest point distance between two sets of point clouds. It has two terms to force the output to be closed to as well as cover the ground truth.

\begin{equation}\label{eq:cd}
  \begin{aligned}
    CD_{loss}(S_1,S_2)= \frac{1}{|S_1|}\sum_{x\in S_1}\min_{y \in S_2}||x-y||_2  \\
    + \frac{1}{|S_2|}\sum_{y\in S_2}\min_{x \in S_1}||y-x||_2
 \end{aligned}
\end{equation}

Our network generates two outputs: Coarse points $P_c$, fine points $P_f$, and each output corresponds to a CD loss with respect to the ground truth.
\begin{equation}\label{eq:loss}
  \begin{aligned}
    Loss = CD_{loss}\:(P_{c},P_{gt}) + \alpha\:CD_{loss}\:(P_{f},P_{gt})
\end{aligned}
\end{equation}

where $\alpha$ is a hyperparameter and $P_{gt}$  are ground truth points. 

\section{Experiments}
This section describes our dataset and the training process first, and then compares our methods with several strong baselines. Afterward, we present the ablation study results, followed by experiment on real-sensor data. Following previous works~\cite{yuan2018pcn,Tchapmi_2019_CVPR,xie2020grnet}, Chamfer
Distance serves as our quantitative evaluation metric for baseline comparison and ablation study.
\subsection{Datasets and Implementation Details}

{\textbf{ShapeNet PCN.} The ShapeNet dataset~\cite{shapenet2015} for point cloud completion is derived from PCN~\cite{yuan2018pcn}, consisting of 30,974 3D models from 8 categories. The ground
truth point clouds containing 16,384 points are uniformly sampled on mesh surfaces. The partial point clouds are generated by randomly sampling 8 camera poses to capture 8 depth images for each CAD model and backing-projecting the 2.5D depth images to 3D, and thus own a variant number of points.}

\textbf{ShapeNet Bottle.} We built a small dataset for ablation study and our future work in manipulator grasping following ShapeNet PCN. It contains 498 bottle-shaped objects such as water bottles, jars, spray bottles, etc. Each object is also used to generate eight partial observations with a flexible number of points and one ground truth point cloud with 9,216 points. 80\% of the models are used for training, 10\% for evaluation and 10\% for testing.

\textbf{YCB Dataset.} It consists of objects of daily life with different shapes, sizes, and textures. The dataset provides mesh models and RGB-D scans of the objects.

We select the following setting based on our implementations and dataset. During constructing the graph representation of the point cloud, each point is connected with its 20 neighbors. KDTree~\cite{muja2009fast} was used for fast retrieval of nearest neighbors. We first generate a coarse point cloud with 1,024 points and link each point with its 5 nearest neighbors. A $4\times4$ grid initialized with $l$ equals to 0.1 is used to sample 16 neighbors for each coarse point. As a result, our model consumes an input with a variable number of points and generates the coarse output with 1,024 points and fine output with 16,384 points. We trained the networks for 200 epochs with a decaying learning rate initialized at $10^{-4}$ and Adam optimizer~\cite{kingma2014adam}, with a batch size of 32 on Four NVIDIA GeForce RTX 2080Ti GPUs. It takes about 78 hours to complete 100 epochs. The average inference time for one instance is 0.075 seconds on a single GPU. 

\begin{table}[]
    \centering
    \begin{tabular}{c|c|c|c|c|c}
    \hline
         Method &FN & GRnet & TopNet& PCN &Ours\\
        \hline
         \# Params & 2.40M & 76.7M  & 3.27M &6.85M & 9.60M\\
    \hline
    \end{tabular}
    \caption{Number of trainable model parameters}
    \label{tab:my_label}
\end{table}

\subsection{Baselines and Performance Comparison}
Here, we compare our model against four baselines that work on point cloud completion.
\begin{enumerate}

    \item\textbf{FoldingNet:} It encodes each point's neighbor information by flattening the covariance matrix and proposed a folding-based decoder that deforms a 2D fixed grid onto the 3D surface of a point cloud~\cite{Yang_2018_CVPR}.

    \item\textbf{TopNet:} It proposes a decoder following a hierarchical rooted tree structure to generate points~\cite{Tchapmi_2019_CVPR}.
    
    \item\textbf{GRnet:} A recent method that uses 3D voxels as intermediate representations for point clouds ~\cite{xie2020grnet}.
    \item\textbf{PCN:} The representative of point completion methods following the coarse-to-fine fashion~\cite{yuan2018pcn}.

\end{enumerate}
Quantitative results in Table~\ref{tab:tab1} and Table~\ref{tab:my_label} indicate our method could achieve better performance over all eight categories while still maintain a reasonable number of parameters. As shown in \figref{fig:exp:compare}, our completion results could recover a more detailed geometry than others. It proves the advantage of our proposed method that extracts both the local information and the global structure information from the observations and the strength of generating various local geometries. 

\subsection{Ablation Study}
In order to investigate our encoder and decoder's performance separately, we train four different neural networks for the ablation study on the ShapeNet Bottle. A $3\times3$ grid is used to generate 9,216 points. We evaluate the performance on the validation dataset every 20 epochs during training.
\begin{enumerate}
\setlength{\itemsep}{0.5pt}
\setlength{\parsep}{0.5pt}
\setlength{\parskip}{0.5pt}
    \item\textbf{Model A}: Removing surface normal and adaptive 3D grid from our decoder (same as PCN's decoder).
    \item \textbf{Model B}: Removing GAT from our encoder (same as PCN's encoder).
\end{enumerate}

\begin{figure}[htbp]
    \centering
    \includegraphics[width=0.4\textwidth]{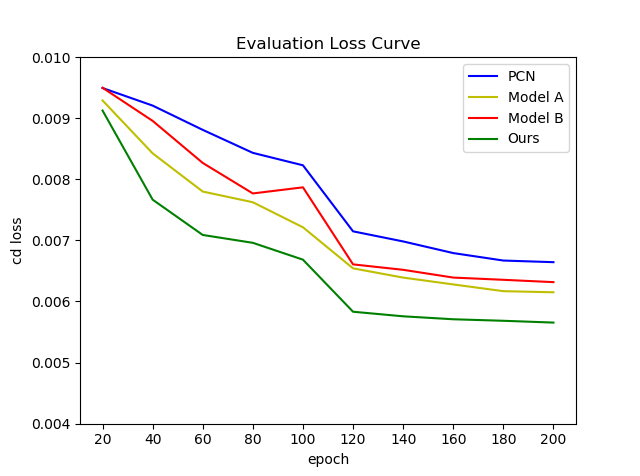}
    \caption{CD loss during evaluation. The vertical axis is the CD loss, and the horizontal axis is the epoch number. We evaluate the performance every 20 epochs on the validation dataset.}
    \label{fig:exp:abla}
\end{figure}

 As shown in \figref{fig:exp:abla}, both our encoder and decoder converge fast and outperform PCN's encoder and decoder. Without GAT in the encoder, the performance is decreased by 8\%; without surface normals and scale factor $\sigma$, the performance is decreased by 10\%. It further indicates our encoder's benefit, which combines local information with global information, and our decoder's power, which utilizes surface normals and adaptive grid sizes.
 
 We also investigated using multiple GAT layers to encode local information, while the test results show no improvement but the deterioration of the performance. We speculate the main reason is that two graph attention layers allow each point to communicate with more neighbors, which will lead it to lose the property of local convolution, and hence be difficult to learn efficiently with the attention mechanism.
\subsection{Experiment on Sensory Data}
\begin{figure}
    \centering
    \includegraphics[width=0.45\textwidth]{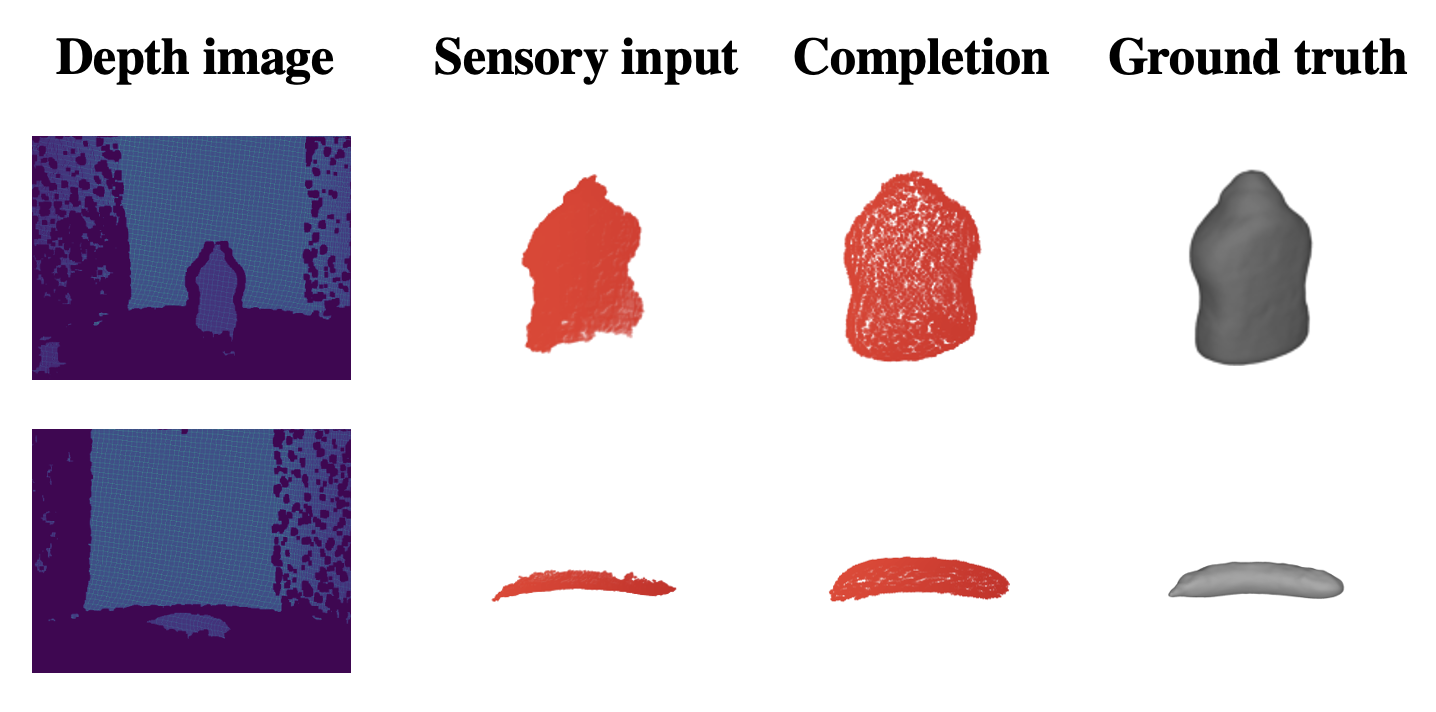}
    \caption{Experimental results on YCB objects, showing completions for a mustard bottle and a banana, respectively. We extract the point clouds (second column) from the raw depth images (first column), and feed them into the model to generate the completions (third column). The ground truth meshes are shown in the last column.}
    \label{fig:my_label}
\end{figure}

We test our model on YCB objects with a reasonable reconstructed mesh (70 in total), as shown in \figref{fig:my_label}. We first build a small simulation dataset using the same strategy mentioned in ShapeNet Bottle. After training our model for 150 epochs, we test it on sensory point clouds extracted from raw registered depth images. Our model achieves an average chamfer distance of $9.9834\times 10^{-3}$ on depth images of the first camera view of each object from the YCB dataset. It is worth noting that the real sensor data is nosier and occasionally suffers from missing geometry features compared to the simulated data.
\section{Discussion}
\subsection{Nearest Neighbor Distance for Each Point}
\begin{figure}
    \centering
    \begin{subfigure}[b]{0.23\textwidth}
        \centering
        \includegraphics[width=\textwidth]{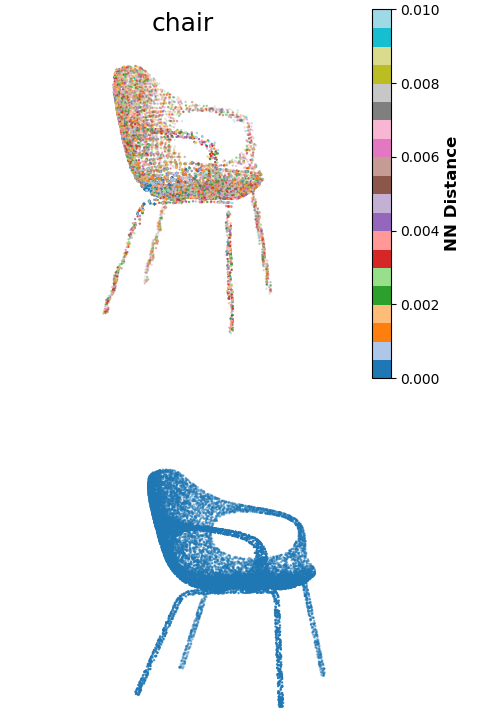}
        \caption{CD 0.0061}
        \label{fig:gull}
    \end{subfigure}
    ~ 
    \begin{subfigure}[b]{0.23\textwidth}
        \includegraphics[width=\textwidth]{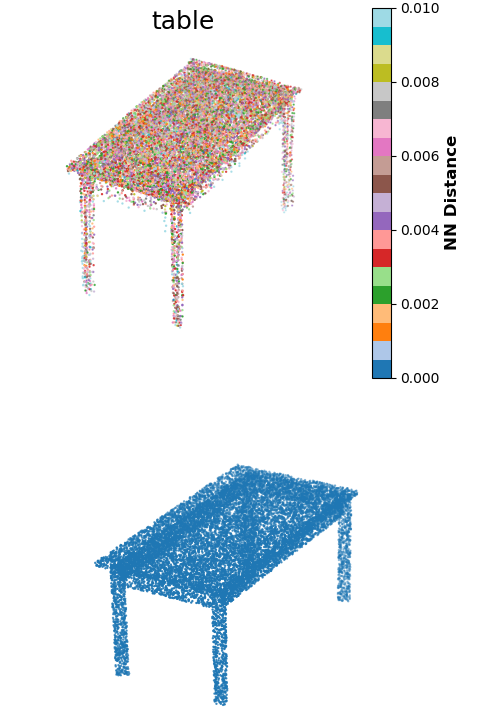}
        \caption{CD 0.0090}
        \label{fig:tiger}
    \end{subfigure}
    \caption{Nearest Neighbor Distance for Each Point. The top row shows our generated outputs, and the bottom row shows the ground truth. The points are rendered by their closest distances to the ground truth.}\label{fig:heatmap}
\end{figure}

Since Chamfer Distance only provides the average performance for the whole point cloud~\cite{xu2019disn}, to analyze our results in detail, we illustrate each point's nearest neighbor distance to the ground truth in \figref{fig:heatmap}. The generated points are rendered by their closest distances to the ground truth. As it suggests, each point maintains a relatively small distance to the closest point in the ground truth.

\subsection{Generalization Ability}

\begin{figure}[htbp]
    \centering
    \small\textbf{{\:\:\:\:Partial \:\:\:\:\:\:\:\:\:\:\:\:\:\:  Coarse \:\:\:\:\:\:\:\:\:\:\:\:\:\:  Fine \:\:\:\:\:\:\:\:\:\:\:  Ground Truth}}\par\medskip
    \includegraphics[width=0.4\textwidth]{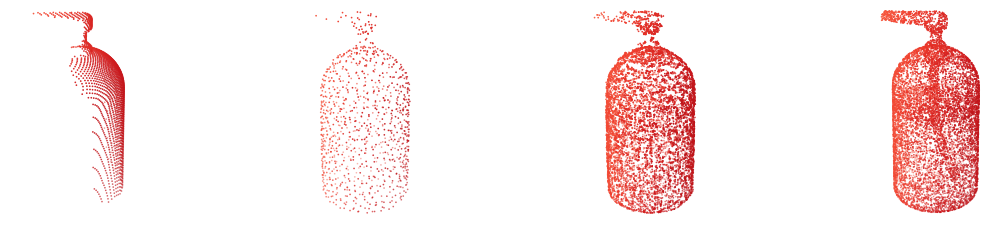} 
    \includegraphics[width=0.4\textwidth]{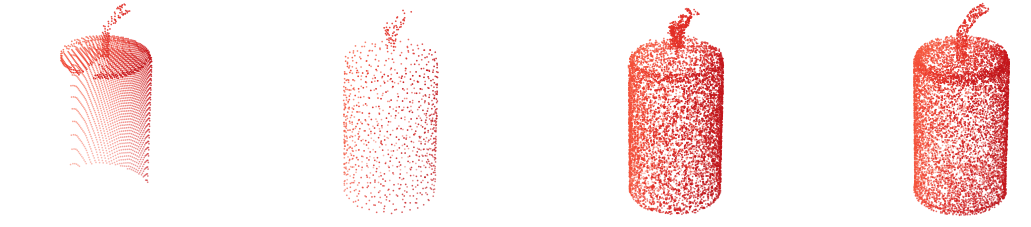} 
    \caption{Illustration of Generalization Ability of GASCN: first, we take a partial point cloud as input and output coarse point clouds, and then do the coarse-to-fine refinement. The top row and the bottom row show the generation process of a shampoo bottle and a coke can with a straw, respectively. Though there is no straw in our training data, our model generates a similar geometry for the unseen part of the partial observation during testing.}
    \label{fig:exp:bottle}
\end{figure}

When testing on the ShapeNet Bottle dataset, we found some interesting results. \figref{fig:exp:bottle} shows our generation process for a shampoo bottle and a coke can with a straw. There is no straw-like object but some shampoo bottles in the training dataset, and our network could still reason out the geometry of the straw, though it resembles the shampoo header to some degree. This phenomenon shows that our model could predict a reasonable shape for the unseen part of the input point cloud, which indicates the generalization ability of our proposed method.

\subsection{Point Cloud Registration}
Many tasks in robotics could benefit from dense complete point clouds. We illustrate its advantage on the point cloud registration. Specifically, we ran ICP~\cite{besl1992method} for two different partial observations from one object and compared its results with their corresponding complete fine points. Our experiment indicated that our generated complete points have fewer errors since they have significant overlaps compared to the partial observations. It is worth noting that our model could enormously increase the utilization of observations in many downstream tasks, especially where both the completions and surface normals are needed.

\begin{figure}
    \centering
    \begin{subfigure}[b]{0.2\textwidth}
        \centering
        \includegraphics[width=\textwidth]{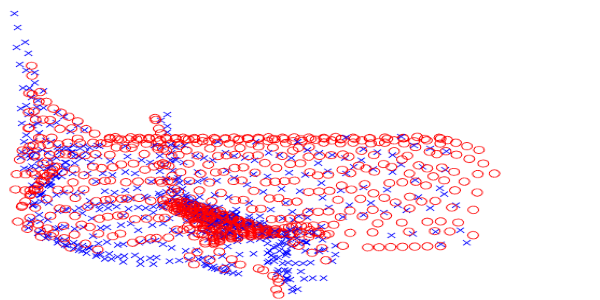}
        \caption{MSE: 0.2915}
        \label{fig:gull}
    \end{subfigure}
    ~ 
    \begin{subfigure}[b]{0.2\textwidth}
        \includegraphics[width=\textwidth]{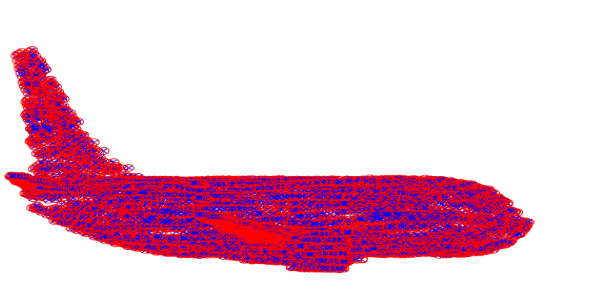}
        \caption{MSE: 0.0657}
        \label{fig:tiger}
    \end{subfigure}
    
    \centering
    \begin{subfigure}[b]{0.2\textwidth}
        \centering
        \includegraphics[width=\textwidth]{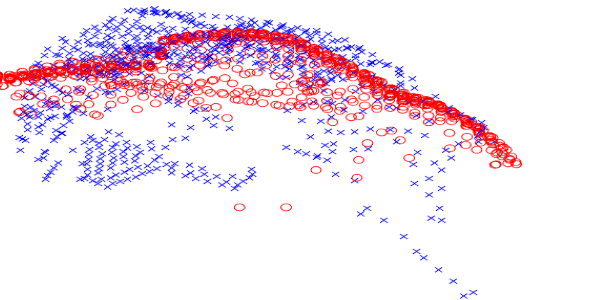}
        \caption{MSE: 1.2068}
        \label{fig:gull}
    \end{subfigure}
    ~ 
    \begin{subfigure}[b]{0.2\textwidth}
        \includegraphics[width=\textwidth]{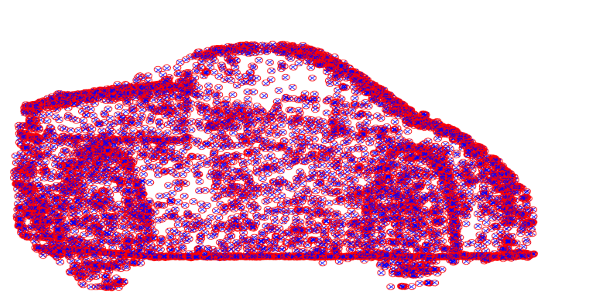}
        \caption{MSE: 0.0836}
        \label{fig:tiger}
    \end{subfigure}
    
    ~ 
    \caption{Improvement on Point Cloud Registration. The left column shows registrations for partial observations, and the right column shows registration for their corresponding complete fine points.}\label{fig:icp}
\end{figure}

\section{Conclusion}
In this work, we studied how to infer the complete 3D geometry from an incomplete observation. Our motivation is to combine local information with global information in our encoder and take the pointwise local property into account during our generation. To achieve the target, we utilized the graph attention layer for encoding local context information and utilized MLPs and max-pooling to aggregate the global structure information. To decode the latent vector properly, we inversed the process of calculating surface normal from points and utilized adaptive 3D grids to obtain neighbors. In our future work, we will investigate combining shape completion with grasp pose detection~\cite{gualtieri2016high} in robotics.

\section*{Acknowledgement}
This work was supported in part by NSF 1724257, NSF 1724191, NSF
1763878, NSF 1750649, and NASA 80NSSC19K1474.
{\small
\bibliographystyle{ieee_fullname}
\bibliography{egbib}
}
\end{document}